\documentclass[letterpaper, 10 pt, conference]{ieeeconf}  %
\IEEEoverridecommandlockouts                              %
\usepackage{amsmath}
\usepackage{lipsum}
\usepackage{graphics}    %
\usepackage{times}       %
\usepackage{amsmath}     %
\usepackage{amssymb}     %
\usepackage{cite}

\newif\ifshortbibnames
\shortbibnamestrue

\usepackage{xcolor}
\usepackage{graphicx}
\usepackage{algorithm}
\usepackage[noend]{algpseudocode}
\usepackage{subcaption}
\usepackage{placeins}    %
\usepackage[flushmargin]{footmisc}
\usepackage{siunitx}
\usepackage[colorinlistoftodos,prependcaption]{todonotes}

\usepackage[most]{tcolorbox}
\usepackage{xcolor}

\definecolor{pastelblue}{RGB}{210,225,245}
\definecolor{pastelblueborder}{RGB}{120,160,210}
\definecolor{puckwrong}{RGB}{235,104,52}
\definecolor{puckwide}{RGB}{42,120,214}
\definecolor{puckadaptive}{RGB}{18,148,135}

\newtcolorbox{resultbox}[1][]{%
  enhanced,
  breakable,
  colback=pastelblue,
  colframe=pastelblueborder,
  boxrule=1.2pt,  
  arc=4pt,
  outer arc=4pt,
  boxrule=0pt,
  left=6pt, right=6pt, top=6pt, bottom=6pt,
  before skip=6pt, after skip=6pt,
  fonttitle=\bfseries,
  title={Result},
  #1
}

\makeatletter
\let\NAT@parse\undefined
\makeatother
\usepackage[hidelinks]{hyperref}
\usepackage[nameinlink]{cleveref}
\Crefname{figure}{Fig.}{Figs.}
\Crefname{table}{Tab.}{Tabs.}
\Crefname{algorithm}{Alg.}{Algs.}

\usepackage[font=small]{caption}

\title{\LARGE \bf Do My Samples Help? Compute Budget Allocation for Real-World Sampling-Based MPC in Contact-Rich Manipulation}

\author{
  Magnus Dierking\textsuperscript{*,1} \and
  João Carvalho\textsuperscript{3,4} \and
  Vince Kurtz\textsuperscript{6} \and
  Jan Peters\textsuperscript{2,3,4,5} \and
  Nils Dengler\textsuperscript{1,5} \and
  Georgia Chalvatzaki\textsuperscript{1,2,5}
}

\begin{document}
\bstctlcite{BSTcontrol}
\maketitle
\thispagestyle{empty} 
\pagestyle{empty}

\let\thefootnote\relax\footnotetext{%
  \textsuperscript{1} PEARL Lab, Dept. of Computer Science, TU Darmstadt, Germany \\
  \textsuperscript{2} Hessian.AI, Darmstadt, Germany \\
  \textsuperscript{3} TU Darmstadt, Institute for Intelligent Autonomous Systems, Dept. of Computer Science, Germany \\
  \textsuperscript{4} German Research Center for AI (DFKI), Research Department: Systems AI for Robot Learning \\
  \textsuperscript{5} Robotics Institute Germany (RIG) \\
  \textsuperscript{6} School of Computing, DePaul University, Chicago, IL, USA 
}
\footnotetext{*Corresponding author: \texttt{magnus.dierking@tu-darmstadt.de}}

\begin{abstract} 
Sampling-based model predictive
control~\mbox{(SMPC)} is increasingly practical for contact-rich manipulation, as
GPU-based physics simulators enable real-time evaluation of large numbers of non-differentiable contact rollouts.
However, this computation budget is finite and can be spent in competing ways, including denser control exploration, representing dynamics or state uncertainty, replanning more frequently, or aligning the simulation better with the physical state.
Yet, the resulting trade-offs are not well understood in real-world sim-to-real control.
In this work, we address this question through a systematic study of how to allocate a fixed computation budget for
SMPC in contact-rich manipulation.
To enable controlled comparisons, we develop a GPU-based framework named Budget-Allocating Model Predictive Control~(BAMPC), which runs on MuJoCo Warp and exposes dynamics uncertainty, state uncertainty, and temporal alignment as independent dimensions of the computation budget while using the same control stack in simulation and on a real robot.
Across three contact-rich manipulation tasks, we find that the best budget allocation depends on the dominant sim-to-real error, with dynamics randomization not universally the most effective one.
\\
Project Page:  \href{https://bampc-framework.github.io}{\textcolor{blue}{bampc-framework.github.io}}

\end{abstract}

\section{Introduction}
\label{sec:intro}

Solving contact-rich manipulation tasks with full trajectory optimization is hard because contact dynamics are non-smooth, and differentiable or analytically tractable models are often inaccurate under real-world contacts.
In this context, Sampling-based Model Predictive Control (SMPC) is a reasonable alternative, as it combines online replanning and gradient-free optimization using forward simulation~\mbox{\cite{honda2026model, jordana2025introduction}}.
Common methods are Model Predictive Path Integral control~(MPPI)~\cite{williams2018information}, the Cross-Entropy Method~(CEM)~\cite{rubinstein2004cem}, and Predictive Sampling~(PS)~\cite{howell2022predictive}.
As recent GPU-based simulators make it possible to evaluate thousands of simulated physics rollouts within a single control cycle~\cite{kurtz2024hydrax, li2025_judo}, SMPC becomes feasible for real-time applications.
However, simulator-based control can fail when rollouts do not match the real world: errors in contact parameters, friction, mass distribution, or collision geometry can produce incorrect contact transitions and object motions, even when the simulator appears physically plausible.

Unlike legged locomotion, where SMPC often exploits simplified normal-force-dominated foot--ground contacts~\mbox{\cite{schramm2025reference, xue2025full}}, contact-rich manipulation depends on both tangential friction and normal forces across diverse geometries, making such approximations invalid, while amplifying model and state uncertainty.
To counter this effect, uncertainty modeling, including dynamics domain randomization (DR), can reduce sensitivity to these errors by evaluating candidate actions across multiple possible system realizations.
However, this robustness comes at a computational cost. Under real-time constraints, samples used to represent dynamics variations cannot simultaneously be used to explore the control distribution more densely, update the controller more frequently, or account for uncertainty in state estimation.

\begin{figure}[t]
    \centering
    \includegraphics[width=\linewidth, trim= 90 65 90 65, clip]{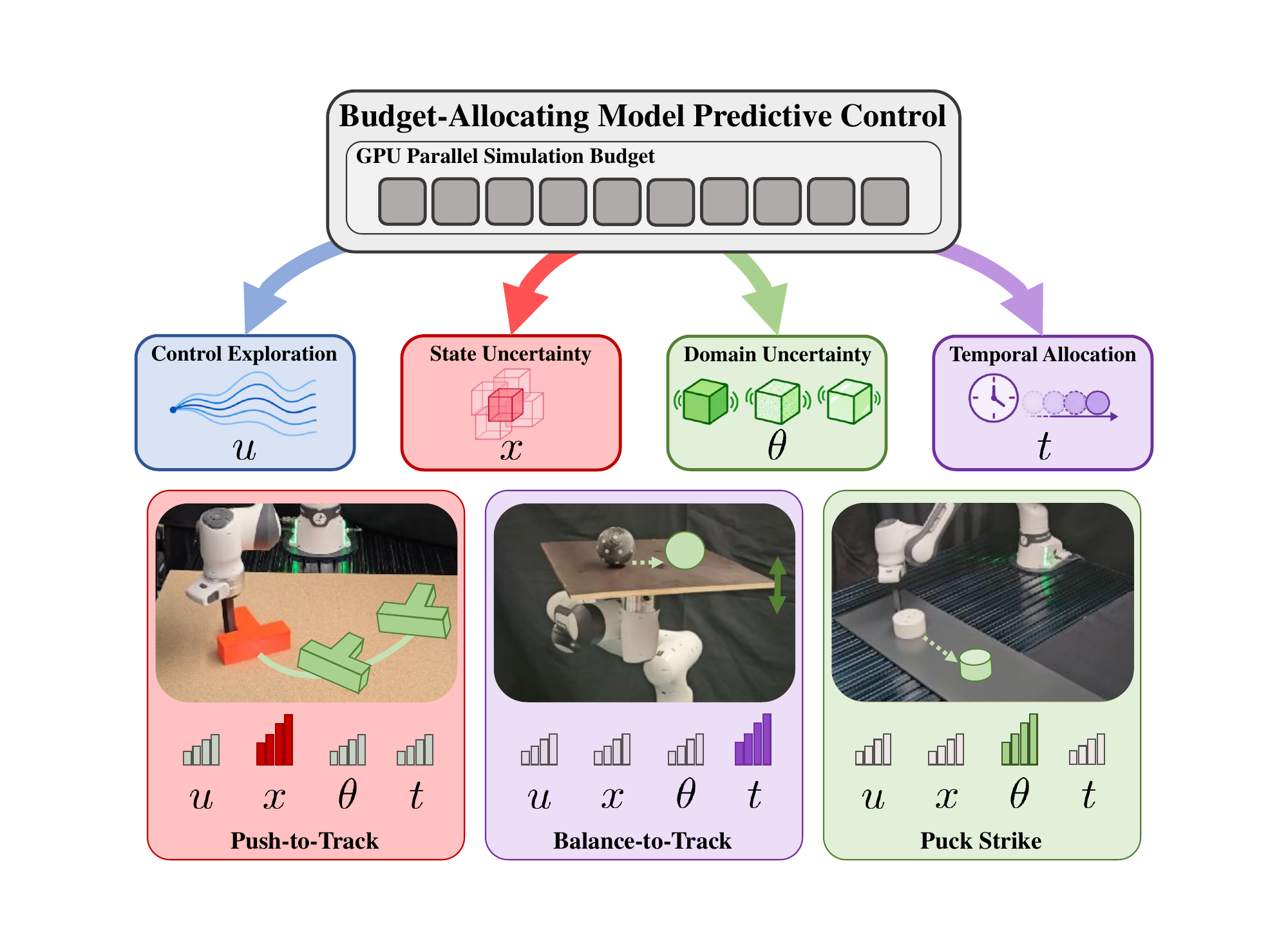}
    \caption{Under real-time constraints,
    SMPC frameworks have a finite simulation budget that must be allocated among competing uses, including control exploration, state uncertainty, dynamics uncertainty, and temporal alignment. The most effective allocation is task-dependent and should reflect the dominant source of sim-to-real error. 
    BAMPC provides a common framework to expose and compare these choices across different tasks.
    }
    \label{fig:pull}
    \vspace{-10px}
\end{figure}

This trade-off has received surprisingly little systematic attention for
SMPC on real contact-rich manipulation systems. 
Prior approaches identify dynamics randomization as a promising direction but do not isolate its effect experimentally \cite{brudermueller2026contact, le2025model}. 
In approaches where dynamics randomization is part of the controller, it is commonly enabled as part of the overall system rather than ablating its effects independently \cite{pezzato2025sampling}. 
Consequently, it remains unclear whether allocating samples to dynamics variation is the most effective use of a fixed real-time sampling budget under sim-to-real discrepancies.
More generally, there is little empirical guidance on how to split a fixed
sampling budget among state uncertainty, dynamics uncertainty, and controller update rate when deploying
SMPC on a real robot.

In this work, we study this sampling-budget allocation problem for sim-to-real
SMPC in contact-rich manipulation. Rather than assuming that uncertainty modeling can improve robustness, we ask: \textit{given a
real-time control constraint and fixed computation
budget, where should additional
samples be spent to obtain the best real-world performance? }

Therefore, our main contribution is a systematic analysis of how a finite simulation budget should be allocated for robust sim-to-real
SMPC in contact-rich manipulation.
We find that the best allocation is determined by which source of sim-to-real error dominates a task, e.g., state, dynamics, or temporal alignment, and demonstrate it on three challenging tasks on a real robot.
For that, we built Budget-Allocating Model Predictive Control~(BAMPC), a novel
GPU-based MPC framework
using MuJoCo Warp, which supports dynamics randomization, state uncertainty, and real-robot integration, including sensing and actuation latency compensation through forward prediction.
The framework executes a large number of rollouts in real time on a single GPU and exhibits favorable scaling as additional simulated worlds are introduced.
Compared to a previous JAX-based implementation~\cite{kurtz2024hydrax} that inspired some architecture choices, it supports substantially larger real-time rollout populations while avoiding compilation-heavy development cycles, enabling faster experimentation and real-world deployment.

\section{Related Work}
\label{sec:related}\noindent
SMPC~\cite{jordana2025introduction, honda2026model} is increasingly used for contact-rich manipulation for two complementary reasons. 
First, sampling-based methods accommodate non-differentiable objectives~\cite{williams2017model, pmlr-v164-bhardwaj22a}, and thus are suited for contact-rich tasks. Second, modern physics simulators provide ready-to-use contact dynamics, substantially reducing the modeling effort required for complex interactions~\cite{li2025_judo, xue2025full, pezzato2025sampling, hurova2025sampling, abraham2020model, hess2025sampling}.\

An open question in this setting is how model and state uncertainty should enter the controller. 
Here, most prior work on uncertainty in
SMPC has focused on the dynamics model. Abraham et al.~\cite{abraham2020model} introduce Ensemble MPPI (EMPPI), which propagates a weighted ensemble of physics-parameter hypotheses through the rollouts and updates their weights online from prediction error. 
This captures uncertainty over latent physical parameters that should decrease as the parameters are identified. 
However, the quantitative benefits of EMPPI are demonstrated primarily in simulation, while the real-world experiments are qualitative and do not establish that maintaining a parameter ensemble improves robustness over a single-model controller.
\mbox{Vahs et al.~\cite{vahs2026parameter}} similarly maintain and refine a particle-based parameter belief to enforce probabilistic safety constraints on a physical quadrotor.
Their gradient-based inference, however, relies on a smooth analytical model with few identifiable inertial parameters.
Furthermore, they assume the state is known, leaving joint parameter and state estimation for future work.
In contrast, we study uncertainty over the current system state, for which maintaining a non-degenerate belief can remain beneficial even with a well-calibrated model.\

Recently, simulation-based work has increasingly adopted per-rollout dynamics randomization, but rarely evaluated its benefit directly~\cite{brudermueller2026contact,hess2025sampling}.
Pezzato et al.~\cite{pezzato2025sampling} randomize quantities such as object mass, friction, and size in simulator-based MPPI to account for model and perception uncertainty, but do not isolate the effect of this randomization.
Similarly, Le et al.~\cite{le2025model} vectorize rollouts across perturbed dynamics models, while focusing their ablations on sampling and exploration rather than on the value of model randomization itself. 
However, their work does not compare model uncertainty against other uncertainty sources under a fixed sampling budget, although this trade-off becomes critical in real-time control, where additional model hypotheses consume compute that could instead be spent on representing state uncertainty or exploring controls. 
In this context, DROP~\cite{li2025drop}, performs real-world cube reorientation with a single model and leaves domain-randomized or risk-sensitive planning to future work due to computational reasons.

\section{Method}
\label{sec:approach}\noindent
This section formalizes the compute-budget allocation problem and describes the
\mbox{SMPC} formulation used to compare different budget allocations.
We consider control exploration, state uncertainty, dynamics uncertainty, and temporal alignment as competing uses of a finite computation budget.
To make these choices comparable, we first formulate uncertainty-aware
SMPC within a common rollout representation and then show how state and dynamics uncertainty map to the same parallel evaluation structure.
Finally, we introduce BAMPC as the
GPU-based execution framework that realizes these allocations while retaining the same planning and deployment stack in simulation and on the real robot.

\begin{figure*}[ht]
    \centering
    \includegraphics[width=\textwidth, trim= 150 40 150 50, clip]{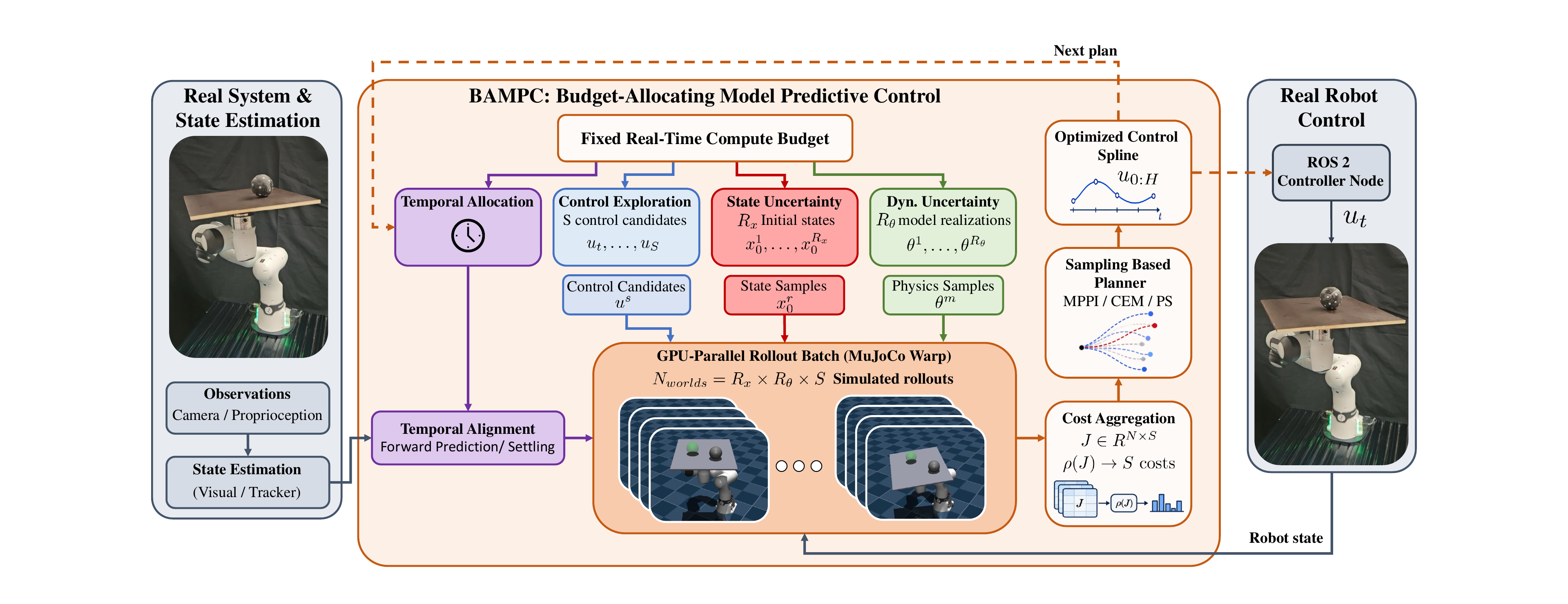}
    \caption{Overview of Budget-Allocating Model Predictive Control (BAMPC). 
    A fixed real-time computation budget is allocated across control exploration, state uncertainty, dynamics uncertainty, and temporal alignment.
    Candidate controls are evaluated under multiple state and physics realizations using
    GPU-based MuJoCo Warp rollouts, aggregated into scalar costs, which are used by the sampling-based planner to produce the optimized spline control points. 
    The same control stack is used in simulation and on the real robot.
    }
    \label{fig:technical}
    \vspace{-5px}
\end{figure*}

\subsection{Sampling-based Model Predictive Control
(SMPC)}
\label{sec:SBMPC}\noindent

SMPC optimizes a finite-horizon control sequence through repeated sampling, rollout evaluation, and distribution updates, as summarized in
\Cref{alg:smpc-bampc}.
At each planning step, from the current state $\mathbf{x}_0\in\mathbb{R}^n$, the planner 
runs up to $K$ optimization steps, where it samples and evaluates $S$ candidate control sequences 
$\mathbf{U}^{(s)}=\{\mathbf{u}_0^{(s)},\ldots,\mathbf{u}_{H-1}^{(s)}\}$ drawn from a distribution $\mathbb{Q}$, where $H$ is the prediction horizon and $\mathbf{u}_t^{(s)}\in\mathbb{R}^m$.
Typically, $\mathbb{Q}(\mathbf{V},\boldsymbol{\Sigma})$ is a Gaussian with mean $\mathbf{V}$ and covariance $\boldsymbol{\Sigma}$.
Each candidate is propagated according to $\mathbf{x}_{t+1}^{(s)}=\mathbf{f}_{\boldsymbol{\theta}}(\mathbf{x}_t^{(s)},\mathbf{u}_t^{(s)})$, where $\boldsymbol{\theta}$ denotes the dynamics parameters.
In our implementation, $\mathbf{f}_{\boldsymbol{\theta}}$ is a MuJoCo Warp physics model~\cite{todorov2012mujoco,Macklin_Warp_A_High-performance_2022}, which allows parallel simulation.
The rollout cost~$J^{(s)}$ combines running and terminal costs
\begin{equation}
    J^{(s)} =
\sum_{t=0}^{H-1}
c\!\left(
    \mathbf{x}_t^{(s)},
    \mathbf{u}_t^{(s)}
\right)
+
c_T\!\left(
    \mathbf{x}_H^{(s)}
\right).
\end{equation}
The planner selects the action sequence $\mathbf{U}^\star$ and applies its first (or a sequence of the next $n$) actions to the system.

We treat MPPI~\cite{williams2018information}, CEM~\cite{rubinstein2004cem}, and
PS~\cite{howell2022predictive} through the same \mbox{\emph{sample--rollout--score--update--select}} interface in
\Cref{alg:smpc-bampc}.
The methods differ in their distribution update and action selection rules.
PS selects the lowest-cost candidate, MPPI performs an importance-weighted update, and CEM updates from an elite subset.
MPPI and CEM apply the first action of the updated nominal sequence, while PS applies the first action of the selected candidate.
We fix $\boldsymbol{\Sigma}$ for all methods and update only $\mathbf{V}$ to avoid premature loss of exploration 
and isolate the effects of budget allocation~\cite{pezzato2025sampling}.
Finally, control sequences are parametrized with splines
to reduce the sampling dimension and increase 
smoothness.

\begin{algorithm}[t]
    \caption{General SMPC and \textcolor{violet}{BAMPC} planning step}
    \label{alg:smpc-bampc}
    \textcolor{violet}{Purple} marks BAMPC extensions.
    \begin{algorithmic}[1]
        \Require Control distribution $\mathbb{Q}$, current state $\mathbf{x}_0$, control samples $S$, optimization iterations $K$
        \Require \textcolor{violet}{Uncertainty realizations $R$, risk measure $\rho$}
        \Statex Indices: $s=1,\ldots,S$\textcolor{violet}{, $r=1,\ldots,R$}.
        \State \textcolor{violet}{Construct state and dynamics realizations $\mathcal{E}^{(r)}$}
        \For{$k=1,\ldots,K$}
            \State $\mathbf{U}^{(s)} = \operatorname{SAMPLE}(\mathbb{Q},S)$
            \State $%
            \mathbf{X}^{(s\textcolor{violet}{,r})} = \operatorname{ROLLOUT}(%
            \mathbf{x}_0\textcolor{violet}{\text{ or }\mathcal{E}^{(r)}},\mathbf{U}^{(s)})$
            \State $%
            J^{(s\textcolor{violet}{,r})} = \operatorname{SCORE}(%
            \mathbf{X}^{(s\textcolor{violet}{,r})},\mathbf{U}^{(s)})$
            \State \textcolor{violet}{$\tilde{J}^{(s)} = \rho(J^{(s,1)},\ldots,J^{(s,R)})$}
            \State $\mathbb{Q} = \operatorname{UPDATE}(\mathbb{Q},\mathbf{U}^{(s)},%
            J^{(s)}\textcolor{violet}{\text{ or }\tilde{J}^{(s)}})$
        \EndFor
        \State $%
        \mathbf{U}^\star = \operatorname{SELECT}(\mathbb{Q},\mathbf{U}^{(s)},%
        J^{(s)}\textcolor{violet}{\text{ or }\tilde{J}^{(s)}})$
    \end{algorithmic}
    \smallskip
\end{algorithm}

\subsection{Uncertainty-Aware Parallel Rollouts}
\label{sec:uncert}\noindent
To represent uncertainty in SMPC, each of the $S$ candidate control
sequences is evaluated under $R$ uncertainty realizations, yielding
$N_{\mathrm{worlds}} = RS$ parallel worlds.
Each realization specifies an initial state $\mathbf{x}_0^{(r)}$
and dynamics parameters $\boldsymbol{\theta}^{(r)}$, while the same
candidate $\mathbf{U}^{(s)}$ is applied across all $R$ realizations. The
resulting rollout costs form
$\mathbf{J} \in \mathbb{R}^{S \times R}$, with entries
$J^{(s,r)} = J(\mathbf{U}^{(s)};\mathbf{x}_0^{(r)},\boldsymbol{\theta}^{(r)})$. 
Thus, the planning time scales approximately as
\begin{align}
    \underbrace{\bigl(N_{\mathrm{settle}} + H\bigr)\,
    \tau(RS)}_{\text{batched rollout}}
    + \tau_{\mathrm{host}}
    + \tau_{\mathrm{forward}}.
\end{align} 
Here, \(N_{\mathrm{settle}}\) denotes the settling steps, \(H\) the horizon, \(\tau(RS)\) the batched rollout time, and \(\tau_{\mathrm{host}}\) and \(\tau_{\mathrm{forward}}\) the host-side and forward-pass times, respectively.
Before the control distribution update, the uncertainty dimension is reduced
independently for each candidate by aggregating the $R$ rollout costs,
yielding ${\tilde{J}^{(s)} = \rho(J^{(s,1)},\ldots,J^{(s,R)})},$ where $\rho$ denotes an aggregation operator, such as the empirical mean.
The planner therefore always receives $S$ scalar candidate costs,
independent of the number or type of uncertainty realizations.

In general, state and dynamics uncertainty differ only in how the $R$ realizations are constructed. 
For state uncertainty, $\mathbf{x}_0^{(r)}$ represents a plausible initial state while $\boldsymbol{\theta}$ remains fixed. 
For dynamics randomization, $\boldsymbol{\theta}^{(r)}$ contains sampled physical parameters such as friction or mass while the initial state is fixed.
Both can also be varied jointly without changing the rollout or cost-reduction structure.
Dynamics parameters are sampled once
and remain fixed over the prediction horizon, although the framework also supports resampling. We use fixed parameters throughout our experiments to isolate their effect and make the resulting behavior easier to interpret.

Temporal alignment differs from ensemble-based uncertainty mechanisms because it uses additional computation to advance or synchronize the simulated state, rather than to evaluate additional state or dynamics hypotheses within the same planning step. The \emph{replanning rate} determines how frequently new feedback is incorporated into the control loop. \emph{Settle steps} advance the simulator before the scored horizon so that transient contact inconsistencies, such as small inter-penetrations, can resolve before planning begins. \emph{Forward prediction} propagates the initial state over the measured planning latency so that the optimized trajectory is aligned with the state expected at execution time. Since all three mechanisms consume additional simulation steps without introducing another rollout dimension, we treat temporal alignment as a separate axis of the computation budget.

\subsection{Budget-Allocating Model Predictive Control (BAMPC)}
\label{sec:framework}\noindent

\begin{figure*}[t]
    \centering
    \begin{subfigure}[t]{0.32\textwidth}
        \centering
        \includegraphics[width=\columnwidth]{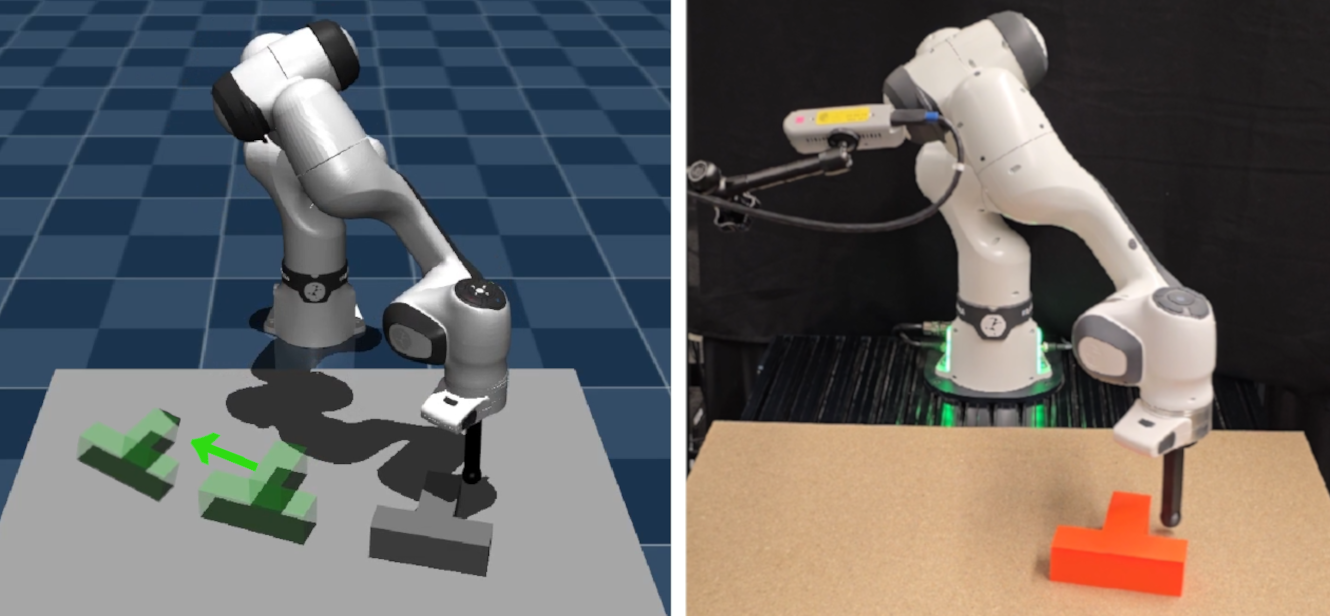}
        \caption{Push-to-Track}
        \label{fig:experimental-overview-push}
    \end{subfigure}\hfill
    \begin{subfigure}[t]{0.32\textwidth}
        \centering
        \includegraphics[width=\columnwidth]{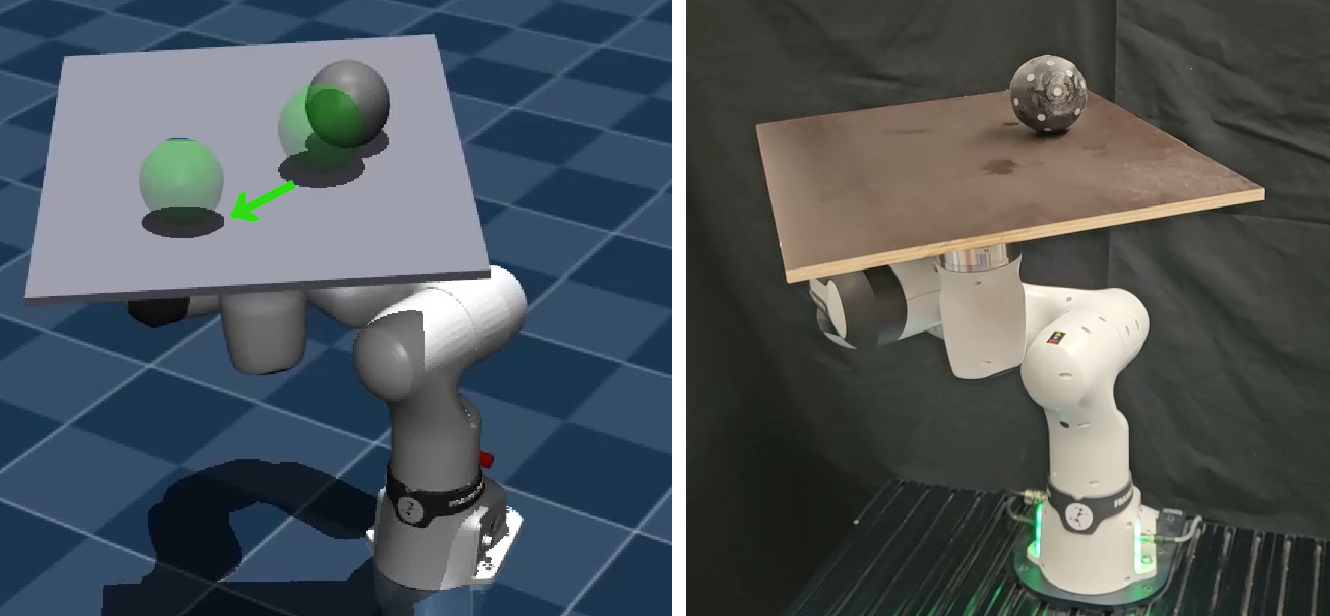}
        \caption{Balance-to-Track}
        \label{fig:experimental-overview-balance}
    \end{subfigure}\hfill
    \begin{subfigure}[t]{0.32\textwidth}
        \centering
        \includegraphics[width=\columnwidth]{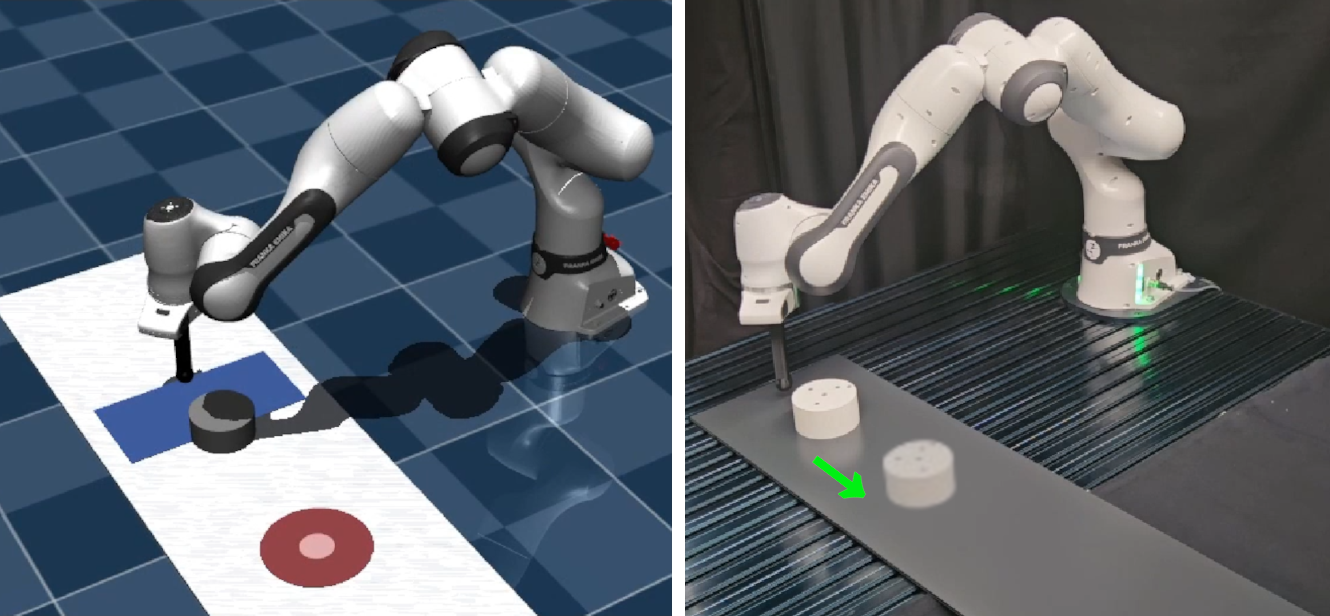}
        \caption{Puck Striking}
        \label{fig:experimental-overview-third-task}
    \end{subfigure}
    \caption{
        Overview of the real-world contact-rich tasks used in our experiments.
    }
    \label{fig:experimental-overview}
    \vspace{-10px}
\end{figure*}

The BAMPC framework is designed to expose competing uses of a finite real-time computation budget within a common planning and deployment stack.
Its architecture is shown in \Cref{fig:technical} and follows the computational structure of
\mbox{SMPC}, where planner updates are comparatively inexpensive while forward simulation dominates the computation and can be parallelized across rollout worlds.
We therefore separate planning from rollout evaluation.
We run sampling and planner updates on the CPU, while trajectory propagation and cost evaluation run in parallel on the GPU using MuJoCo Warp.
At each planning step, the CPU-side planner generates $S$ candidate control sequences, and the GPU evaluates the corresponding $N_{\mathrm{worlds}} = RS$ rollout worlds across state and dynamics realizations.
BAMPC evaluates each candidate across uncertainty realizations and aggregates their costs before updating the control distribution.
The complete $H$-step simulation, including cost accumulation, is captured once as a CUDA graph and replayed at every planning iteration.
This minimizes repeated CPU--GPU coordination and limits data transfer back to the planner.

\Cref{alg:smpc-bampc} summarizes one BAMPC planning step, where expressions in \textcolor{violet}{purple} mark BAMPC alternatives or additional steps compared to SMPC.
Here, $\mathcal{E}=\{(\mathbf{x}_0^{(r)},\boldsymbol{\theta}^{(r)})\}_{r=1}^{R}$ contains the state and dynamics realizations, and $\operatorname{ROLLOUT}$ evaluates all $RS$ worlds in parallel.
The allocation of $(S,R,K)$ is configured to fit the available computation budget, and $\rho$ is the empirical mean in our experiments, but other risk measures can be used~\cite{esteban2026surprising}.

Because planning is expensive and robot execution requires commands at higher and more regular rates, BAMPC decouples the planning rate from the control rate.
Consequently, the planner updates the optimized trajectory whenever a new solution is available, while the controller continuously evaluates the most recent spline at the higher control rate.
This enables smooth high-rate actuation minimizing the impact of planning latency on time-critical command execution, while preserving the same planner interface in simulation and on hardware.
BAMPC also accounts explicitly for computation spent on temporal alignment between the simulated rollout state and the physical system.
Before the scored prediction horizon begins, \emph{settle steps} advance the simulator to resolve transient contact inconsistencies in the initialized state.
In addition, \emph{forward prediction} propagates the latest state estimate over the measured planning and actuation latency so that optimization starts from the state expected when the command becomes active.
Both operations use simulation steps and hence compete with control exploration and uncertainty resolution for compute budget.

\section{Experimental Evaluation}
\label{sec:exp}\noindent

In our experiments, we use a Franka Emika FR3 with low-level joint velocity control
 and represent manipulated objects as free rigid bodies with
unconstrained six-degree-of-freedom motion.
For fast collision checking, robot links are approximated by capsules, and no
artificial joints or constraints are introduced to suppress object motion.
Such simplifications increase simulation speed, but retain rolling, sliding, and contact-transition effects that contribute to sim-to-real
discrepancies.
For the contact configurations in our experiments, we require a reduced simulation timestep and an increased solver iteration count (in contrast to~\cite{pezzato2025sampling,xue2025full}).

Our experiments examine how a real-time computation budget is allocated under
different dominant sources of sim-to-real error.
We use three contact-rich manipulation tasks
(see \Cref{fig:experimental-overview}) that span markedly different regimes:
\emph{Push-to-Track} (\Cref{subsec:push-to-track}) is quasi-static planar
pushing under repeated visual occlusion;
\emph{Balance-to-Track} (\Cref{subsec:balance-to-track}) demands highly
dynamic and reactive motion to control a rolling object;
and \emph{Puck Striking} (\Cref{sec:strike}) is governed by a single brief
non-prehensile impact.
For each task, we test all three allocation axes under the same budget and report the results for the dominant axis.

\subsection{Experimental Procedure}
\label{sec:experimental-protocol}\noindent
All comparisons use task-specific real-time compute budgets measured on an
NVIDIA RTX~5090 (\Cref{sec:scaling}).
We treat control exploration as the reference allocation, since increasing the
number of candidate controls is the standard use of additional SMPC
computation, and reallocate part of this budget to state uncertainty, dynamics
uncertainty, or temporal alignment.
For the state and dynamics axes, the rollout population
$N_{\mathrm{worlds}} = RS$ is held fixed, so raising the number of uncertainty
realizations $R$ directly reduces the number of control candidates $S$.
Temporal alignment respect the same real-time constraint but may alter
$N_{\mathrm{worlds}}$ or the planning frequency, which we state per
condition.
In each task, all other planner and simulation settings are fixed across
conditions.
Initial states are drawn from frozen scenario sets, and each condition is
repeated across multiple states and runs because MuJoCo Warp rollouts are
non-deterministic~\cite{mjwarp_docs}.

\subsection{Framework Runtime Performance}
\label{sec:scaling}\noindent

\begin{figure*}[t]
    \centering
    \includegraphics[width=0.9\textwidth, trim=10 18 0 5, clip]
    {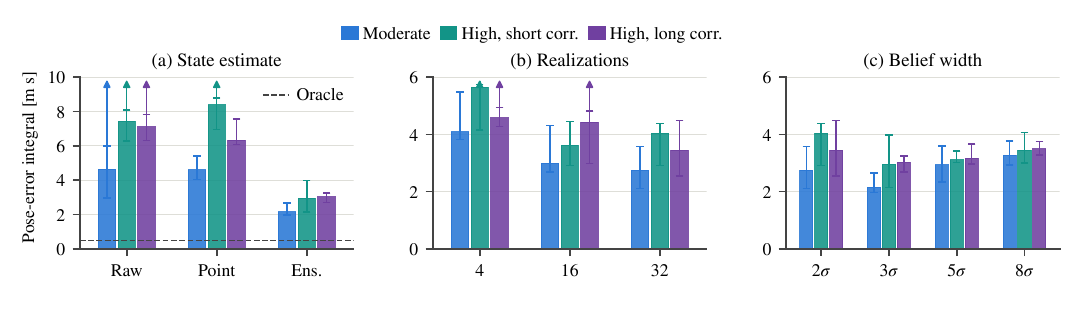}
\vspace{-3px}
\caption{
    Push-to-Track ablations in simulation.
    (a) The particle-based state ensemble (Ens.) consistently improves over raw and point-state representations across the tested sensing conditions.
    (b) Increasing the number of state realizations can improve performance.
    (c) An intermediate belief width performs best.
    The variance of the baselines Raw and Point in (a) is caused by failure cases in which the robot pushes the object out of reach, a failure not observed with the state ensemble.
    All experiments use $13500$ samples at \SI{10}{\hertz}.}
    \label{fig:experiment-push-sim}
    \vspace{-10px}
\end{figure*}

Before studying budget allocation, we characterize how many parallel rollouts
BAMPC can evaluate within a real-time compute budget.
Throughout our experiments, we freeze all numerical settings and planning horizons per task.
Runtime then grows approximately linearly with the number of simulated physics
steps, but sublinearly with the number of parallel worlds, so large rollout
populations can be evaluated without a proportional increase in planning time.

We compare BAMPC against Hydrax~\cite{kurtz2024hydrax}, a
closely related and broadly used GPU-based SMPC implementation using JAX and
MJX.
At identical horizon, timestep, solver, and sampling settings in the pushing
setup, BAMPC requires approximately $\SI{34.3}{\milli\second}$ for
$N_{\mathrm{worlds}}=256$, compared with $\SI{164.2}{\milli\second}$ for Hydrax,
corresponding to a
\mbox{\textbf{\SI[number-unit-product={}]{4.8}{\times} speedup}}.
The difference increases to \SI[number-unit-product={}]{18.8}{\times} at
$N_{\mathrm{worlds}}=3010$, the largest batch size supported by Hydrax in our
comparison.

To convey the resulting scale, the pushing setup sustains on the order of
$\num{14000}$ parallel worlds at \SI{10}{\hertz}, the balancing setup roughly
$\num{3600}$ at \SI{20}{\hertz}, and the longer-horizon striking setup
approximately $\num{4100}$ at \SI{10}{\hertz}.
These values are measured in simulation, without settle steps or real-world
overhead, and therefore upper-bound what is available on hardware.
The budgets used in the following experiments are fixed below these limits and
stated per task.

\subsection{Push-to-Track}
\label{subsec:push-to-track}

\subsubsection{\textbf{Task Design}}
Push-T is a common task for planar contact-rich manipulation, combining contact switching and sensitivity to the object pose. 
The standard task terminates at a fixed target~\cite{chi2025diffusion}, and contact and occlusion depend strongly on the initial configuration. Therefore we introduce \emph{Push-to-Track}, where a moving target forces sustained pushing, repeated re-alignment, and recurring visual occlusion.
In simulation, we model the object as a rigid body with friction between contacting surfaces.
In the real-world, its pose is estimated with FoundationPose++~\cite{Wenhao_Yan_and_Jie_Chu_FoundationPose_2025}. 
Controls are sampled in task space $(x,y)$.
The planning rate is \SI{10}{\hertz}, since increasing it to \SI{20}{\hertz}
provided no measurable benefit in simulation, while lower rates lead to visibly less smooth robot motions. Hence, we keep temporal alignment fixed and compare control exploration against state uncertainty.
We report CEM and PS results, which provide distinct update behaviors.
MPPI becomes competitive when its effective sample size approaches one, where its update behaves similarly to PS.

\begin{figure}[t]
    \centering
    \includegraphics[width=0.95\columnwidth, trim=12 5 0 0, clip]
    {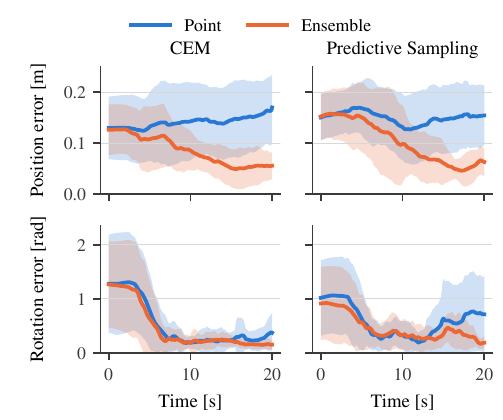}
    \caption{
        Real-world Push-to-Track results with point-state and
    state-ensemble planning for CEM and PS. 
    Curves show means and standard
    deviation over $20$ trials. The ensemble
    reduces position error for both planners and improves rotational robustness
    for PS under repeated visual occlusion. 
    }
    \label{fig:experiment-tracking-real}
    \vspace{-10px}
\end{figure}

\subsubsection{\textbf{Allocating Samples to State Uncertainty}}
Under repeated occlusion, pose errors can change the predicted contact interaction and ranking of candidate controls.
To address this, we track the object pose with a Kalman filter.
Online covariance adaptation was unstable due to highly correlated noise, hence, we calibrated it during a stationary
phase
and then freeze it.
We sample an ensemble of $R=16$ particles from the filter, and evaluate $S=600$ control sequences for each particle state.
Trading off $R$ and $S$ balances state representation and control exploration.
As a baseline, we compare against the Kalman filter mean estimate (point-state) ($R=1$) with $S=9600$ control samples.

The simulation results in \Cref{fig:experiment-push-sim} show the benefit from planning with state uncertainty.
The ensemble achieves lower error than both the raw and point-state estimates (\Cref{fig:experiment-push-sim}a).
Increasing the number of state particles from 4 to 16 or 32~(\Cref{fig:experiment-push-sim}b) generally improves performance, with diminishing returns at larger ensembles.
A belief-width test~(\Cref{fig:experiment-push-sim}c) further shows that more uncertainty is not always beneficial.
An intermediate width around $3\sigma$ performs best, while narrower and substantially wider beliefs increase the error.
Finally, in our experimentation, mean cost aggregation consistently outperformed Conditional Value at Risk~(CVaR) and inverse-CVaR reductions.
Hence, for real-world evaluation, we use a moderate state ensemble with mean aggregation.

The real-world results in \Cref{fig:experiment-tracking-real} confirm the simulation findings.
With point estimates, after initially aligning the T block, PS tends to make aggressive corrections and can overshoot when pose estimates drift during occlusion, whereas CEM updates more conservatively and can remain trapped in local alignment corrections without reducing translational error.
Representing multiple plausible states improves both behaviors despite reducing the number of control candidates.

\subsubsection{\textbf{Alternative Budget Allocations}}
\begin{figure}[t]
    \centering
    \includegraphics[width=0.95\columnwidth, trim=15 10 0 19, clip]
    {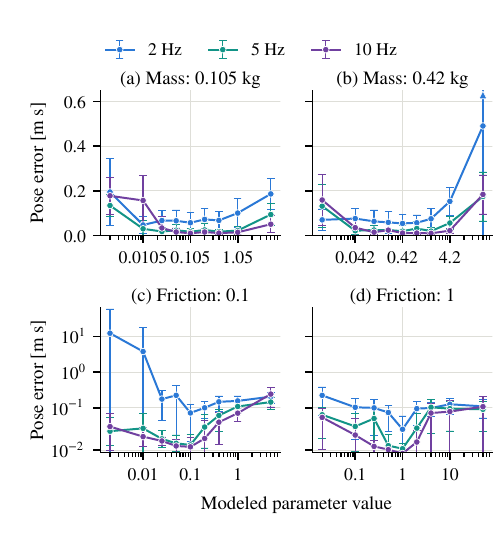}
    \caption{Sensitivity of Push-to-Track to mass and friction mismatch at
    different
    replanning rates.
    The panel titles give the
    true
    parameter values and the horizontal axes show the values used
    for planning. 
    At \SI{10}{\hertz}, tracking is robust 
    for a broad range around the true value, while extreme mismatch and lower
    replanning rates increase error. 
    Shown are the mean and 
    standard
    deviation over four initial states and four repeats. $75000/28000/13500$ samples used at $2/5/10$~\si{\hertz}.}
    \label{fig:experiment-tracking-dr-sensitivity}
    \vspace{-10px}
\end{figure}

To determine whether the benefit arises from using state uncertainty, we compare against alternative uses of samples. 
Here, using a point-state estimate ($R=1$), we allocate the rollout budget to exploring $S$ distinct control samples.
As shown in \Cref{fig:experiment-tracking-real}, the worse real-world results show that additional control exploration does not compensate for uncertainty in the estimated object pose.
Therefore, we test whether compute would instead be better spent on dynamics uncertainty.
\Cref{fig:experiment-tracking-dr-sensitivity} shows results when the modeled mass and friction are changed around two ground-truth parameter values and three replanning rates. 
With \SI{10}{\hertz}, tracking remains robust over a broad range around the true parameters, with substantial degradation under extreme mismatch or at lower replanning rates.

Together, these experiments identify state uncertainty as the effective allocation axis for Push-to-Track. Under the same real-time compute constraint, additional control exploration does not recover the performance of the state ensemble, and plausible dynamics mismatch has comparatively little influence at the operating replanning rate.

\subsection{Balance-to-Track}
\label{subsec:balance-to-track}\noindent

\begin{figure}[t]
    \centering
    \includegraphics[width=0.95\linewidth, trim= 0 33 0 35, clip]{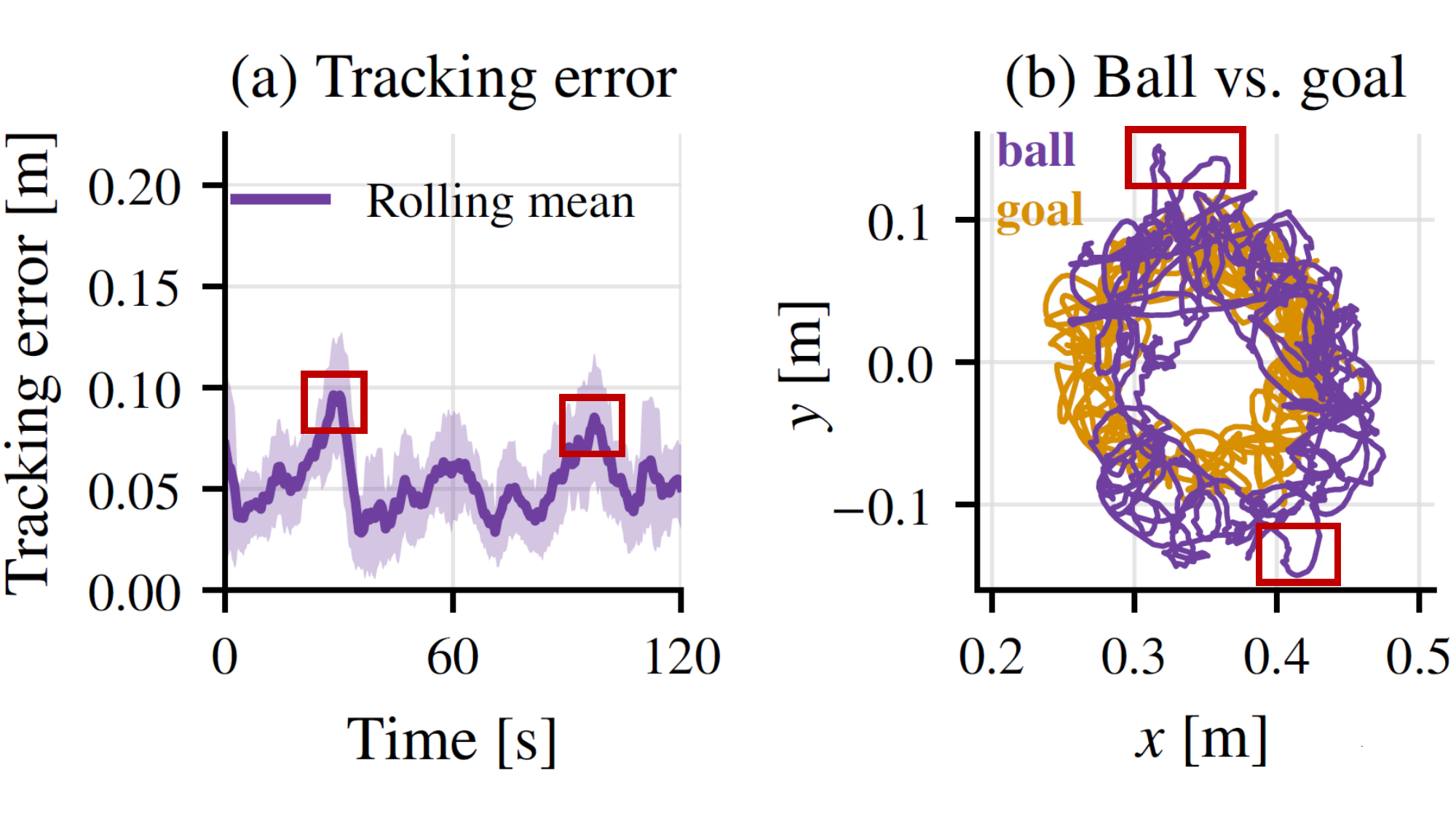}
    \caption{
        Real-world tracking error of Balance-to-Track.
        (a) 5-s rolling mean of the ball-to-goal tracking error. Shading denotes one standard deviation over 5 runs.
        (b) Corresponding ball and goal trajectories in the $xy$ plate plane.
        $1200$ samples used at \SI{20}{\hertz}.}
    \label{fig:experiment-balancing}
    \vspace{-10px}
\end{figure}

\subsubsection{\textbf{Task Design}}
Balance-to-Track complements Push-to-Track by replacing repeated occlusion and multi-contact pushing with fast rolling dynamics and high-dimensional joint-space control. 
A ball is placed on a $39 \times$\SI{50}{\centi\meter} plate 
and must remain balanced while tracking a spherical target of \SI{8}{\centi\meter} diameter that continuously moves along a circular trajectory with \SI{8}{\centi\meter} radius around the plate center.
A motion-capture system tracks the ball's motion.
The plate has no rim or other constraint preventing the ball from falling, and its non-rectangular geometry alternates between regions with little and limited recovery space.

\subsubsection{\textbf{Allocating Samples to Temporal Alignment}}

Small errors in the initialized ball rolling state can cause the simulated ball to move at a different speed than the real ball, leading the planner to correct too early or too late. Hence, we use \emph{settle steps} before the scored horizon to let the contact state evolve into a dynamically consistent rolling state. These steps consume simulation time and directly trade temporal alignment against control exploration.

During real-world deployment, settle steps strongly affect task performance. With 0-5 steps, the simulated ball rolls too slowly and the planner reacts too late. Around 10--15 steps, simulated and physical motion are sufficiently aligned for reliable tracking, while at 25 steps the simulated ball advances too far and again introduces mismatch. Further results are provided in the supplemental material.
With 12 settle steps, and planning at \SI{20}{\hertz}, the controller balances and tracks the moving target reliably for 
2 minutes, with bounded tracking error throughout the episode as shown in \Cref{fig:experiment-balancing}, while also remaining stable under additional perturbations.
The longest continuous run lasted \SI{2}{\minute}\,\SI{45}{\second}.
Simulation confirms our findings, since providing the full rolling state yields reliable control, while omitting angular velocity reproduces the delayed response observed on hardware.

\begin{figure}[t]
  \centering
  \includegraphics[width=0.95\columnwidth, trim=15 10 0 19, clip]{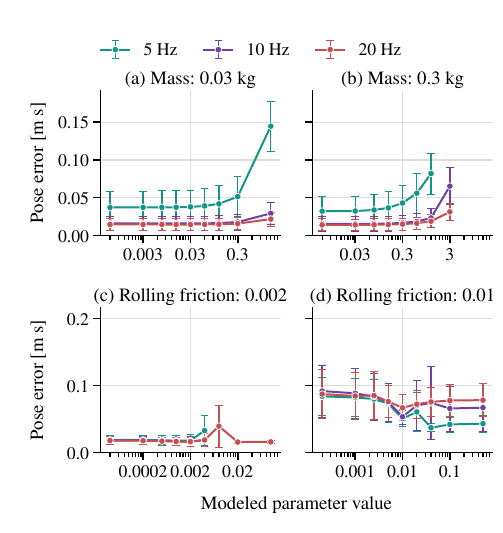}
\caption{Balance-to-Track sensitivity to mass (top) and rolling-friction (bottom) mismatch at two parameter values and different replanning rates. 
Curves show mean accumulated pose error and one standard deviation over four seeds and four repeats.
(c) In the low-friction range, rolling is nearly non-dissipative and underestimating friction has little effect, whereas overestimating it introduces artificial dissipation and a corresponding model mismatch. This range is also readily identifiable from the observed rolling behavior. Missing mass results are caused by numerical instability, while large rolling-friction values are outside the physically plausible range for the rigid sphere. $18500/9000/2800$ samples used at $5/10/20$~\si{\hertz}.}
  \label{fig:experiment-balancing-dr-sensitivity}
  \vspace{-10px}
\end{figure}

\subsubsection{\textbf{Alternative Budget Allocations}}
We test whether we can recover the same performance by allocating samples differently.
Without settling, the saved computation can be used for more control candidates and enable replanning at $\approx\SI{40}{\hertz}$. 
This setting does not lead to stable tracking, since the mismatch in the initialized rolling motion remains.
Therefore, more frequent feedback cannot compensate for poor temporal alignment.
We also tested simply scaling the estimated angular velocity, but it did not reproduce the benefit of settling steps, indicating that the improvement comes from allowing the simulated contact state to evolve consistently with the simulation.

State uncertainty is less relevant than in Push-to-Track because the ball remains visible and its position is estimated reliably. Hence, we test whether dynamics uncertainty warrants additional compute. In \Cref{fig:experiment-balancing-dr-sensitivity} (a, b), we observe that mass has little influence at \SI{20}{\hertz}, while varying standard friction shows no measurable effect. 
In \Cref{fig:experiment-balancing-dr-sensitivity} (c, d), rolling friction shows some sensitivity in the low-dissipation range, but this effect is strongly one-sided.
Underestimating the already small dissipation has little effect, whereas overestimating it introduces artificial damping and a clear
mismatch.
However, system identification can easily identify this behavior, so it does not require online dynamics randomization.

These results identify temporal alignment as the dominant budget allocation axis for Balance-to-Track. Additional control exploration, state realizations, or dynamics variation do not compensate for a poorly initialized rolling ball state.

\subsection{Puck Striking}
\label{sec:strike}

\begin{figure}[t]
    \centering
    \includegraphics[width=0.95\linewidth, trim=6 5 5 2, clip]{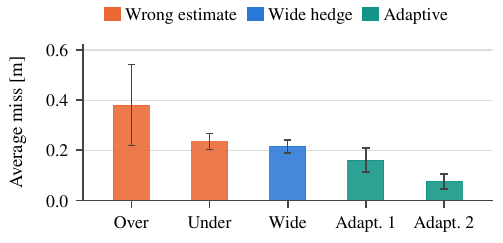}
    \caption{
        Puck Striking real-world results under friction mismatch.
        Bars show the mean target and one standard deviation of final position error over 10 trials.
    Planning with an \textcolor{puckwrong}{over-/underestimated} friction value produces large target errors, while evaluating candidate actions across a \textcolor{puckwide}{wide} range of friction values reduces the average miss.
    \textcolor{puckadaptive}{Adaptively} refining the friction range further improves accuracy over successive updates.
    Planning done with CEM at 10~\si{\hertz} with $(1, 4000)$ or $(10, 400)$ ($R,S$) samples.
    }
    \label{fig:experiment-puck}
    \vspace{-10px}
\end{figure}

\subsubsection{\textbf{Task Design}}
In Push-to-Track and Balance-to-Track, moderate mass and friction mismatch had little effect at the operating replanning rates, a trend we also observed in several experimental tabletop settings.
We attribute this to MPC's short execution interval, where feedback corrects prediction error before it can accumulate.
Puck Striking tests a setting where dynamics mismatch directly affects task performance.
The robot strikes a motion-captured puck from a launch region toward a target. After impact, the puck evolves open-loop until it stops. 
Because the puck's position depends heavily on mass and friction, an inaccurate dynamics model can predict a substantially different stopping location from the ground truth, and the longer the post-impact horizon, the more this difference accumulates.

\subsubsection{\textbf{Allocating Samples to Dynamics Uncertainty}}
To confirm this sensitivity, we run the task with low
($\mu=0.01$) and high ($\mu=0.8$)
friction estimates
and observe the expected under-/overshooting.
Evaluating candidate actions across this range of $1/\mu$-spaced friction values, while splitting samples between control candidates and dynamics hypotheses, improves performance only marginally.

BAMPC supports adapting this split without rebuilding the rollout pipeline: since changing the dynamics realizations only modifies the corresponding rollout parameters, the split between control candidates and dynamics hypotheses can be adjusted online without recapturing the CUDA graph.
After an initial computation with the full range, we score each domain's predicted puck trajectory against the observed one by comparing prediction and observation at matched instants near the end of the horizon.
The resulting per-domain error induces a softmax posterior over friction hypotheses.
Once this posterior collapses, 
we change the sample allocation by
halving the number of domains $R$, doubling the control samples $S$, and resampling friction values over a range centered at the posterior mean.
The successive iterations in \Cref{fig:experiment-puck} show significant improvement with this procedure, which uses the friction hypotheses as a simple tool for online system identification.

\subsubsection{\textbf{Alternative Budget Allocations}}
Since the puck is never occluded during the motion, we observed no advantage of using state uncertainty.
Temporal alignment is similarly less limiting: a small number of settle steps ($2$) suffices to remove initialization artifacts, and more frequent replanning provided no additional benefit during experimentation. 
Therefore, Puck Striking provides the complementary case to the previous tasks:
when dynamics mismatch affects the long-term outcome,
allocating compute to multiple dynamics realizations can become beneficial.

\section{Discussion}
\label{sec_discussion}\noindent

Across the three case studies, useful computation allocations differed with the dominant mismatch and the opportunity for feedback correction.
State ensembles improved Push-to-Track under pose uncertainty, while temporal alignment improved Balance-to-Track under inconsistent rolling states.
In Puck Striking, where feedback cannot correct motion after impact, a broad friction ensemble helped only marginally, while refining its hypotheses from observations improved accuracy and freed samples for control exploration.

These differences highlight the distinction between parameter uncertainty and the sensitivity of predicted outcomes to its variation.
An ensemble may be useful when both are substantial, as a poorly known but inconsequential parameter offers little reason to reduce control exploration, while an influential but accurately known parameter may require only a single estimate.
Even when both conditions hold, feedback may limit the benefit of representing parameter uncertainty.
Our mass and friction tests illustrate this dependence on feedback, with performance degrading mainly under large mismatch or slower replanning (\Cref{fig:experiment-tracking-dr-sensitivity,fig:experiment-balancing-dr-sensitivity}).

For tasks with similar error characteristics, we therefore recommend starting with sufficient control exploration and reallocating compute when a measurable failure mode motivates it.
Perturbing plausible error sources can help assess whether additional state or dynamics hypotheses, or improved temporal alignment, could address that failure.

\section{Conclusion}
\label{sec:conclusion}

This work studied how a fixed sampling budget should be allocated for real-world sampling-based model predictive control~(SMPC) in contact-rich manipulation. %
To enable controlled comparisons, we introduced Budget-Allocating Model Predictive Control~(BAMPC), a GPU-based framework that exposes control exploration, state and dynamics uncertainty, and temporal alignment as competing uses of computation. Across three tasks, allocating compute to the dominant sim-to-real error proved more effective than simply increasing control exploration. State ensembles improved pushing under occlusion, temporal alignment enabled reliable balancing by correcting inconsistent physics predictions, and dynamics ensembles improved puck striking when friction mismatch accumulates over time. These results show that domain randomization should be treated as a task-dependent allocation rather than a default, with computation directed toward the error source that limits real-world performance.

In practical deployments, however, state, dynamics, and temporal errors can coexist and interact. For example, inaccurate dynamics may affect the state propagation used for temporal alignment, linking the treatment of multiple error sources. Extending compute allocation and BAMPC to such combined mismatches is therefore a natural next step, together with adaptive budget allocation and evaluation on higher-dimensional systems beyond tabletop manipulation.
\vspace{-5px}
\section{Acknowledgments}
\label{sec:acknowledgements}
This work is supported by ERC grant SIREN (101163933).
Funded by the European Union. Views and opinions expressed are however those of the author(s) only and do not necessarily reflect those of the European Union or the European Research Council Executive Agency. Neither the European Union nor the granting authority can be held responsible for them. We also acknowledge support from the Alfred Krupp Foundation.
Generative AI was used for textual refinement and proofreading in the preparation of this work. Claude Opus 5 (Anthropic), via Claude Code, provided programming and debugging assistance; all code was reviewed
and validated by the authors.

\bibliographystyle{IEEEtran}

\bibliography{bibliography}

\end{document}